\documentclass[accepted]{uai2023} 

\usepackage[american]{babel}
\usepackage{microtype}
\usepackage{multirow}
\usepackage{enumitem}
\usepackage{bm,amsmath,amssymb}

\usepackage{natbib} 
\usepackage{mathtools} 
\usepackage{booktabs} 
\usepackage{color}
\usepackage{colortbl}
\usepackage{tikz} 

\usetikzlibrary{bayesnet}
\usetikzlibrary{arrows}
\usetikzlibrary{backgrounds}

\graphicspath{{figures/}}

\newcommand\blfootnote[1]{%
  \begingroup
  \renewcommand\thefootnote{}\footnote{#1}%
  \addtocounter{footnote}{-1}%
  \endgroup
}

\newcommand\Item[1][]{%
  \ifx\relax#1\relax  \item \else \item[#1] \fi
  \abovedisplayskip=0pt\abovedisplayshortskip=0pt~\vspace*{-\baselineskip}}

\newcommand*{\T}{^{\mkern-1.2mu\mathsf{T}}}     
\newcommand*{\I}{^{\mkern-1.2mu\mathsf{-1}}}    
\newcommand{\nr}[1]{_{\mkern+2.0mu\mathsf{#1}}} 
\newcommand*{\s}{^{\mkern-1.2mu(\ell)}}    
\newcommand*{\su}[1]{^{\mkern-1.2mu(#1)}}    

\newcommand{\mb}[1]{\mathbf{#1}}
\newcommand{\bs}[1]{\bm{#1}}

\newcommand{\lse}{\(\log\)-\(\mathrm{sum}\)-\(\exp\)}   

\newcommand{\bow}{\textit{bag-of-words}}
\newcommand{\adam}{\textsc{adam }}

\newcommand{\EN}{\textsc{EN}}
\newcommand{\DE}{\textsc{DE}}
\newcommand{\FR}{\textsc{FR}}
\newcommand{\IT}{\textsc{IT}}
\newcommand{\ES}{\textsc{ES}}
\newcommand{\RU}{\textsc{RU}}
\newcommand{\JA}{\textsc{JA}}
\newcommand{\ZH}{\textsc{ZH}}

\newcommand{\GLC}{\textsc{GLC}}
\newcommand{\GLCU}{\textsc{GLCU}}

\newcommand{\MCLRU}{\textsc{MCLRU}}

\newcommand{\myfrac}[2]{     %
  \setbox0\hbox{$#1$}        
  \dimen0=\wd0               
  \setbox1\hbox{$#2$}        
  \dimen1=\wd1               
  \ifdim\wd0<\wd1            
  \dfrac{\hfill#1}{#2}       
  \else                      
  \dfrac{#1}{\hfill#2}       
  \fi
}

\title{A Bayesian Multilingual Document Model for Zero-shot Topic Identification and Discovery}

\author[1]{\href{mailto:<kesiraju@fit.vutbr.cz>}{Santosh Kesiraju}}
\author[*]{Sangeet Sagar}
\author[1]{Luk\'{a}\v{s} Burget}
\author[*]{Ond\v{r}ej Glembek}
\author[1]{J\'{a}n \v{C}ernock\'{y}}
\author[2]{Suryakanth V Gangashetty}
\affil[1]{%
    Speech@FIT, Brno University of Technology, Czechia 
}
\affil[2]{%
	KL University, India
}

\begin{document}
\maketitle

\begin{abstract}
In this paper, we present a Bayesian multilingual document model for learning language-independent document embeddings. The model is an extension of BaySMM~\citep{Kesiraju:2020:BaySMM} to the multilingual scenario. It learns to represent the document embeddings in the form of Gaussian distributions, thereby encoding the uncertainty in its covariance. We propagate the learned uncertainties through linear classifiers that benefit in zero-shot cross-lingual topic identification. Our experiments on 17 languages show that the proposed multilingual Bayesian document model performs competitively, when compared to other systems based on large-scale neural networks (LASER, XLM-R, mUSE) on 8 high-resource languages, and outperforms these systems on 9 mid-resource languages. We revisit cross-lingual topic identification in zero-shot settings by taking a deeper dive into current datasets, baseline systems and the languages covered. We identify shortcomings in the existing evaluation protocol (MLDoc dataset), and propose a robust alternative scheme, while also extending the cross-lingual experimental setup to 17 languages. Finally, we consolidate the observations from all our experiments, and discuss points that can potentially benefit the future research works in applications relying on cross-lingual transfers. 
\end{abstract}

\section{Introduction}\label{sec:intro}
\blfootnote{\textsuperscript{*}While at Speech@FIT BUT}
The zero-shot cross-lingual topic identification (ID) or document classification aims to classify documents from target languages using a classifier trained on examples from one or more source language(s). This is mainly useful in scenarios where the data from target language(s) have little or no labels to train an in-language classifier. Such tasks also have real-world applications~\citep{strassel:2016:lorelei}.
Furthermore, the cross-lingual transfer experiments can also help to analyse and test the capabilities of an underlying multilingual language model.

The common approach is to first train a multilingual language model that aims to capture the semantic relations of words in context, independent of the language~\citep{Ammar:2016:MMWE,Mikel:2019:MMS,Unicoder:2019,Alexis:2020:XLR,Feng:2020:LaBSE}. Such a multilingual model can then later be either (i) fine-tuned for classification~\citep{Siddhant:2019:MMTE} task using labelled examples from source language(s), or (ii) used to extract low-dimensional embeddings (representations) for documents from both source and target languages~\citep{Reimers:2020:Making}; the embeddings from source language(s) together with annotated labels are then used for training a light-weight \textit{independent} classifier for cross-lingual topic ID, which is then used to classify embeddings from target languages.
The former approach relying on fine-tuning is not efficient as it would require to keep a copy of the entire multilingual model (or bunch of adapter layers) for every source language, and every down-stream task. The latter approach of extracting language-agnostic document (sentence) embeddings is more practical as it would require only one model, and several light-weight downstream classifiers.  This paper entirely focuses on models, experiments and analysis related to the latter scheme relying on language agnostic document embeddings, followed by a light-weight classifier.

\subsection{Training multilingual models}
Majority, if not all, of the recent works in multilingual representations for
cross-lingual transfers have relied on training LSTMs~\citep{Schwenk:2017:MNMT,Mikel:2019:MMS} or transformers~\citep{Wu:2019:mBERT,Alexis:2020:XLR} with
huge amounts of data (e.g. 227M - 25B sentences)~\citep{Wu:2019:mBERT,Siddhant:2019:MMTE}. The
pre-training objectives vary depending on the kind of resources used for training such models. In brief, some require parallel translations of sentences across multiple languages, while others rely on bilingual dictionaries~\citep{Ammar:2016:MMWE} or just monolingual texts covering several languages. Training these large multilingual language models requires enormous computational resources~\citep{Emma:2019:Energy}, there is a need for alternatives that are computationally efficient. A majority of the large multilingual models share a vocabulary of sub-word units across several (100) \textit{seed} languages. One needs to take care so that all the languages are equally represented in the shared (sub-)word vocabulary to avoid any language bias from the high resource languages. Additionally, such a design choice makes it challenging to extend these models to newer languages having a different orthography. A fair comparison among these language models is nearly impossible as no two models are trained on exactly the same data. The comparisons are only on the downstream tasks while ignoring the affect of the quality and quantity of pre-training data.  When training on large amounts of web-data it is possible that some of the down-stream data could have been seen during pre-training. Extensive survey on the aforementioned models/approaches can be found in \citep{Ruder:2019:CL_Survey,Sumanth:2021:Primer}.

  In contrast to the neural models, there is also work on classical multilingual topic models~\citep{Mimno:2009:Poly,Yang:2019:MTM}, which are suitable for topic ID and document clustering. While these models are budget-friendly in terms of computation, the  downstream evaluation datasets and tasks~\citep{Schwenk:2018:MLDoc,Kakawani:2020:IndicNLP,Ruder:2020:XTREME} do not overlap between neural and classical models, hence it is difficult to ascertain the advantages of the latter over former.

  \subsection{Contributions of the paper}
  \begin{itemize}
    \item We propose a simple, yet efficient multilingual Bayesian (MBay) model for learning language-agnostic document (sentence) embeddings, that enables to train robust downstream linear classifiers for zero-shot cross-lingual topic ID. 
    \item The proposed model can be easily extended to newer languages without requiring to re-train from scratch (continual learning), while constraining only on a subset of existing parameters, thus making it computation-budget-friendly. Our implementation is public\footnote{\texttt{\url{https://github.com/BUTSpeechFIT/BaySMM}}}.
    \item We re-visit the zero-shot cross-lingual document classification task, and make the following contributions: (i) we identify the shortcomings in evaluation, and propose a robust alternative, (ii) we setup and evaluate zero-shot transfer systems on a new set of 9 languages from IndicNLP suite~\citep{Kakawani:2020:IndicNLP}, in addition to the existing 8 from MLDoc~\citep{Schwenk:2018:MLDoc}, (iii) we benchmark several pre-trained models, and also the proposed model on the revised datasets covering 17 languages (128 transfer directions\footnote{$9$ languages from IndicNLP news articles dataset resulting in $9\times8=72$, and $8$ languages from MLDoc resulting in $8 \times 7=56$ transfer directions $(72+56=128)$.}, and (iv) we provide an in depth analysis of the downstream classification systems, that can best make use of the language-agnostic document (sentence) embeddings from various models.
  \end{itemize}
  \begin{figure}[!t]
    \centering
      \scalebox{0.8}{
        \begin{tikzpicture}
          [
          observed/.style={minimum size=1cm,circle,draw=gray!50,fill=gray!40},
          unobserved/.style={minimum size=1cm,circle,draw},
          hyper/.style={minimum size=1pt,circle,fill=black,scale=0.5},
          post/.style={->,>=stealth',semithick},
          ]
          \node (x-i) [observed] at (0,0) {$\mb{x}\s_{d}$};
          \node (w) [unobserved] at (-2.5,0) {$\mb{w}_d$};
          \node (w-mean) [hyper] [label=left:$\mb{\mu}$] at (-4.3,0.7) {};
          \node (w-std) [hyper] [label=left:$\lambda$] at (-4.3,-0.7) {};
          \node (m) [hyper] [label=above:$\mb{m}\s$] at (-0.5,2) {};
          \node (T) [hyper] [label=above:$\mb{T}\s$] at (0.5,2) {};
          %
          \path
          (w) edge [post] (x-i)
          (w-mean) edge [post] (w)
          (w-std) edge [post] (w)
          (m) edge [post] (x-i)
          (T) edge [post] (x-i)
          ;
          %
          \node [draw,fit=(x-i) (m) (T), inner sep=0.8cm, yshift=0.1cm]
          (plate-token) {};
          \node [above right] at (0.6, -1.1) {$L$};
          \node [draw,fit=(x-i) (w), inner sep=1.1cm, xshift=0.15cm, yshift=-0.35cm]
          (plate-context) {};
          \node [above right] at (1, -1.9) {$D$};
      \end{tikzpicture}}
      \caption{Graphical representation of the proposed multilingual Bayesian model,
        where \(L\) represents number of languages and \(D\) denotes number of
        \(L\)-way parallel documents (translations). \(\{\mb{m}\s,
        \mb{T}\s\}\,\forall \ell\)
        are document-independent, language-specific model parameters, whereas
        \(\mb{w}_d\) is document-specific but language-independent random
        variable (embedding), and $\mb{x}\s_d$ is the observed vector of word counts representing document $d$ from language $\ell$.} \label{fig:graphical_model}
    \end{figure}

  \section{MBay: Multilingual Bayesian Model}
  \label{sec:model}
  %
  Like majority of the probabilistic topic and document
  models~\citep{Blei:2012:PTM,Miao:2016:NVI}, the presented model also relies on
  \bow~representation of documents. Let \(V\s\) represent the vocabulary size in
  language \(\ell \in \mathcal{M}\), where $L=|\mathcal{M}|$ denotes the number of languages. Let \(\{\mb{m}\s,\mb{T}\s\}\, \forall \, \ell\)
  represent the language-specific model parameters, where $\mb{T}\s$ is a
  low-rank matrix of size $V{\s} \times K$ $(K \ll V\s)$ that defines the
  subspace of document specific unigram distributions, and $\mb{m}{\s} \in
  \mathbb{R}^{V\s}$ represents bias or offset.
  The multilingual model assumes that
  the $L$-way parallel data (translations of \bow) are generated according to the
  following process:



  First, a \(K\)-dimensional  language-independent,
  document-specific embedding is sampled from an isotropic Gaussian
  distribution with precision $\lambda$
  %
  \begin{equation}
    \mb{w}_{d} \sim \mathcal{N}(\mb{w} \mid \mb{0}, (\lambda\mb{I})^{\I})
    \label{eq:prior}.
  \end{equation}
  $\mb{w}_d$ can be interpreted as vector representing higher-level semantic
  concepts (such as topic) of a document, independent of any language.
  For each language  \(\ell \in \mathcal{M}\), a vector of word counts
  \(\mb{x}\s_{d}\) is generated by the following two steps:
  The document-specific unigram distribution $\bs{\phi}\s_{d}$ is
  computed using the language-specific parameters
  \begin{align}
    \bm{\phi}\s_{d} = \mathrm{softmax}(\mb{m}\s + \mb{T}\s \, \mb{w}_d),
  \end{align}
  and the vector of word counts \(\mb{x}\s_{d}\) is sampled
  $\mb{x}\s_{d} \sim \mathrm{Multinomial}(\bm{\phi}\s_{d} ,\, N\s_{d})$,
  where $N\s_{d}$ are the number of word tokens in document $d$.
  %
  $\mb{x}\su{1} \ldots \mb{x}\su{L}$ represent $L$-way parallel
  \bow~statistics.
  The above steps describe the generative process of the proposed multilingual
  document model. However, in reality, we do not generate any data, instead
  we invert the generative process: given the training (observed) data
  \(\mb{x}\s_d \, \forall \, \ell \in \mathcal{M},\, \forall \, d=1 \ldots D\), we
  estimate the language-specific model parameters \(\{\mb{m}\s, \mb{T}\s\}\)
  and also the posterior distributions of language-independent document
  embeddings \(p(\mb{w}_d | \mb{x}_d\su{1} \ldots \mb{x}_d\su{L}) \, \forall \,
  d\). Moreover, given an unseen document \(\mb{x}\s_u\) from any of the $L$
  languages, we infer the corresponding posterior distribution of
  the document embedding \(p(\mb{w}_u \mid \mb{x}\s_u)\). Note that such a
  posterior distribution also carries the uncertainty about the estimate.

  Although we describe the model assuming $L$-way parallel data, in practice the
  model can be trained with parallel text (translations) between language pairs
  (bi-texts) covering all the $L$ languages.
  \subsection{Variational Bayes training}
  \label{ssec:elbo}
  The proposed model is trained using the variational Bayes framework, i.e.,
  we approximate the intractable true posterior with the variational
  distribution $q(\mb{w}_d) = \mathcal{N}(\mb{w}_d \mid \bs{\nu}_d, \,
  \mathrm{diag}({\bm{\gamma}_d})\I)$ and optimize the evidence lower-bound~\citep{Bishop:2006:PRML}.
  Further, we use Monte Carlo samples via the re-parametrization
  trick~\citep{Kingma:2014:AEVB,Rezende:2014:SBP}
  to approximate the expectation over \lse~($\log$ normalizer) term which appears
  in the
  lower-bound~\citep{Kesiraju:2020:BaySMM}. The resulting
  lower-bound for a single set of \(L\)-way parallel documents is
  \begin{multline}
    \label{eq:elbo_doc}
    \mathcal{L}(q_d) \approx \sum_{\forall \ell \in \mathcal{M}}
    \sum_{i=1}^{V\s}
    x\s_{di} \,\Big[\,(m\s_{i}\, + \,\mb{t}\s_{i}\,\bs{\nu}_d) \\
    -\frac{1}{R} \sum_{r=1}^{R} \log \Big(\sum_{j=1}^{V} \exp\{m\s_{j} \,+\,
    \mb{t}\s_{j} \,g(\bs{\epsilon}_{dr})\}\Big) \Big] \\
    - D_{\mathrm{KL}}(q_d \mid\mid p) ,
  \end{multline}
  where \(D_{\mathrm{KL}}(q_d \mid\mid p)\) is the Kullback-Leibler divergence
  from variational distribution $q(\mb{w})$ to the
  prior~\eqref{eq:prior} and, $g(\bs{\epsilon}_{dr}) = \bs{\nu} + \bs{\gamma}
  \odot
  \tilde{\bs{\epsilon}}_{dr},$ with $\tilde{\bs{\epsilon}}_{dr} \sim
  \mathcal{N}(\bs{\epsilon} \mid \mb{0}, \mb{I})$. 
  $R$ are the number of Monte Carlo samples used for empirically
  approximating the expectation over \lse.

  The complete lower-bound is just the summation over all the documents.
  Additionally, we use \(\ell_2\) regularization term with weight \(\omega\)
  for language-specific model parameters \(\{\mb{T}\s\} \, \forall\, \ell\).
  Thus, the final objective is
  \begin{equation}
    \label{eq:objective}
    \mathcal{L} = \sum_{d=1}^{D} \mathcal{L}(q_d) \,-\, \omega \sum_{\forall \ell
      \in \mathcal{M}} \,\sum_{i=1}^{V\s} \,||\,\mb{t}\s_{i}\,||_{\nr{2}}\,.
  \end{equation}
  In practice, we follow batch-wise stochastic optimization of
  \eqref{eq:objective} using \adam~\citep{Kingma:2014:Adam}. For a batch of
  documents $d \in \mathcal{B}$ covering a subset of languages
  $\mathcal{M}_B \subseteq \mathcal{M}$,
  we update the all model parameters $\{\mb{m}\s,\mb{T}\s\}\, \forall \, \ell \in
  \mathcal{M}_B$
  and the variational posterior distribution of document embeddings
  $q(\mb{w}_d) \, \forall \, d \in \mathcal{B}$.

  \subsection{Extending to newer languages}
  \label{ssec:extend}
  Since the model uses language-specific parameters and vocabulary, it is
  possible to extend the model to a new set languages (denoted by
  $\bar{\mathcal{M}}$) without re-training from scratch. The necessary
  conditions are that every new language ($\bar{\ell}$) should have parallel text with at least one other language from $\mathcal{M} \cup \bar{\mathcal{M}}$ subject to the constraint that there exists at least one parallel pair between $\mathcal{M}$ and $\bar{\mathcal{M}}$.
  This can be seen as continual learning, and requires only to learn the parameters corresponding to the newer
  languages $\{\mb{m}\bar{\s}, \mb{T}\bar{\s}\}\, \forall \, \bar{\ell} \in \bar{\mathcal{M}}$. It also means that the performance on existing seed languages is unaffected with the addition of newer languages.
  In this paper, we show the results from experiments where we start with a seed model
  covering 6 languages, which is then extended to 11 newer languages. Similar approaches are also explored for multilingual neural machine translation~\citep{Alex:2021:Cont}.

  \subsection{Inferring embeddings}
  \label{ssec:xtr}
  Given a bag-of-word statistics from an unseen document from any of the \(\ell \in \mathcal{M} \cup \bar{\mathcal{M}}\)
  languages, we can infer (extract) the corresponding document embedding
  along with its uncertainty. This is done by keeping the language-specific
  model parameters \(\{\mb{m}\s, \mb{T}\s\}\) constant, and iteratively
  optimizing the objective in~\eqref{eq:elbo_doc} with respect to the
  parameters of the variational distribution. In the resulting variational
  posterior \(q(\mb{w}) =
  \mathcal{N}(\mb{w}
  \mid \bs{\nu}, \,\mathrm{diag}({\bm{\gamma}})\I) \), the mean \(\bs{\nu}\)
  represents the (most likely) document embedding, and variance
  \(\mathrm{diag}(\bs{\gamma})\I\) encodes the uncertainty around the
  mean \(\bs{\nu}\). Since all the documents and language-specific model
  parameters are independent (Fig.~\ref{fig:graphical_model}), inferring the
  embeddings can be parallelized
  and is computationally cheaper.

  %
  %
  \section{Classification exploiting uncertainties}
  \label{sec:clf}
  In a typical setting where we have only point estimates of embeddings,
  all the embeddings are considered equally important by a classifier. This
  may not be true all the time. For example, shorter and documents with many rare
  words can result in poor estimates of the embeddings; which can affect parameters of the
  classifier during training, and also the performance during prediction.
  Additionally, there might be noise while projecting embeddings from multiple languages into the same semantically aligned latent space. The proposed model yields document embeddings represented by Gaussian distributions, with the uncertainty about the embedding encoded in the
  covariance. These uncertainties are specific to each example and can be seen
  as \textit{heteroscedastic aleatoric} uncertainties~\citep{Kendall:2017:Uncert}.
  We present two linear
  classifiers that can exploit this uncertainty.
  The first one is the generative Gaussian linear classifier with uncertainty
  (\GLCU)~\citep{Kesiraju:2020:BaySMM}. The second one is the discriminative
  multi-class logistic regression with uncertainty (\MCLRU).

  \subsection{Generative classifier}
  \label{ssec:clf1}
  In generative classifiers, the posterior
  probability of class label ($\mathcal{C}_k$) given a feature vector (embedding)
  $\mb{w}$ is computed from the joint distribution
  %
  \begin{equation}
    p(\mathcal{C}_k \mid \mb{w}) = \frac{p_{\theta}(\mb{w} \mid
      \mathcal{C}_k)\,p(\mathcal{C}_k)}{\sum_j
      p_{\theta}(\mb{w} \mid \mathcal{C}_j)\,p(\mathcal{C}_j)}
  \end{equation}
  where, \(p_{\theta}(\mb{w} \mid \mathcal{C}_k)\) is the likelihood function
  parameterized by \(\theta\), and \(p(\mathcal{C}_k)\) is the class prior. In
  generative classifiers, the likelihood function is assumed to have a
  known parametric form (e.g. Gaussian, Multinomial).
  For Gaussian linear classifier (\GLC), the likelihood function is
  $p_{\theta}(\bs{w} \mid \mathcal{C}_k) = \mathcal{N}(\bs{w} \mid \bs{\mu}_k,
  \mb{S}\I)$, where $\bs{w}$ is the input feature (point estimate of the
  embedding),
  $\bs{\mu}_k$ is the mean of class $\mathcal{C}_k$, and $\mb{S}$ is the
  precision matrix shared across all the classes.

  Given that the input features come in the form of Gaussian
  distributions, i.e., \(q(\mb{w}) = \mathcal{N}(\mb{w}
  \mid \bs{\nu}, \,\mathrm{diag}({\bm{\gamma}})\I) \), we can integrate out
  (exploit) the uncertainty in the input while evaluating the likelihood
  function. In the case of GLC, where the likelihood
  function is also Gaussian, the expected likelihood has an
  analytical form:
  \begin{equation}
    \mathbb{E}_q[p_{\theta}(\mb{w} \mid \mathcal{C}_k)]
    = \mathcal{N}(\bs{\nu} \mid
    \bs{\mu}_k, \mb{S}\I + \mathrm{diag}(\bs{\gamma})\I).   \label{eq:lh_glcu}
  \end{equation}
  \GLC~with likelihood function replaced by~\eqref{eq:lh_glcu} is called \GLCU.
  Both are essentially the same classifiers, i.e., they have the
  same assumptions about the underlying data and hence the same model parameters.
  The only difference lies in the evaluation of likelihood function.

  \subsection{Discriminative classifier}
  \label{ssec:clf2}
  For discriminative classifier such as multi-class logistic regression (LR), the posterior probability of a class ($\mathcal{C}_k$) given an input feature vector $\mb{w}$ is
  \begin{equation}
    \label{eq:clf_softmax}
    p(\mathcal{C}_k \mid \mb{w}) = \frac{\exp\{\mb{h}\T_k \mb{w} + b_k\}}{\sum_j
      \exp\{\mb{h}\T_j \mb{w} + b_j\}},
  \end{equation}
  where $\{b_k,\, \mb{h}_k\}\, \forall\, k$ are the parameters of the classifier.
  Unlike in GLC, we cannot analytically compute the expectation
  over~\eqref{eq:clf_softmax} with-respect-to the input embeddings (Gaussian distributions). Instead we approximate the expectation using Monte Carlo samples~\citep{Xiao:2019:AAAI}:
  \begin{equation}
    p(\mathcal{C}_k \mid \mb{w})
    \approx \frac{1}{M} \sum_{m=1}^{M} \frac{\exp\{\mb{h}\T_k \bs{\varepsilon}_m +
      b_k\}}{\sum_j \exp\{\mb{h}\T_j \bs{\varepsilon}_m + b_j\}},
    \label{eq:clf_softmax_mc}
  \end{equation}
  $\bs{\varepsilon}_m \sim q(\mb{w}) \, \forall \,
  m.$ Eq.~\eqref{eq:clf_softmax_mc} represents the posterior probability
  computation for logistic regression with uncertainty (LRU).

  Theoretically, given the true uncertainties in the training examples,
  GLCU and LRU can better estimate the model parameters of the classifier. Similarly, it can also exploit the uncertainties in the test examples during prediction.
  However, in our case, the uncertainties are estimated using the Bayesian multilingual document model as described in Section~\ref{ssec:xtr}. The underlying assumption here is
  that uncertainties extracted using the model are close enough to the true uncertainties as expected by the classifiers, which is empirically supported through our experimental results presented in Section~\ref{sec:res}.




  %
  \section{Experimental setup}
  \label{sec:exp}
  This section presents the details on data for multilingual training of MBay model and dataset preparation for downstream classification (topic ID) task. We also discuss the details of various pre-trained multilingual models and downstream classifiers that are used in our experiments.
  \subsection{Data for multilingual training}
  \label{ssec:dataset}
  The following datasets were used for training the proposed MBay model. Europarl(v7) \citep{Koehn:2005:Europarl}, UNPC(v1) \citep{Ziemski:2016:UNPC}, MultiUN(v1) \citep{Eisele:2010:MultiUN}, Global-Voices(v2018q4) \citep{Tiedemann:2012:OPUS}, News-Commentary(v16) \citep{WMT:2021}, CVIT(PIBv1.3, MKB) \citep{Siripragada:2020:CVIT}, Samanantar(indic2indic) \citep{Samanantar:2022}, Japanese-English Wikipedia, and CCAligned(EN-JA) \citep{CCAligned:2020}. The total number of sentences used are 17.89M covering 17 languages. All the words were lower-cased and punctuation was stripped. Further,
  words that do not occur in at least two sentences were removed. We used scikit-learn~\citep{scikit-learn} for pre-processing. More details are given in Appendix A.

  \subsection{Dataset preparation for topic ID}
  \label{ssec:data_prep}
  The original MLDoc corpus was prepared~\citep{Schwenk:2018:MLDoc} in order
  to have a standard training, development (dev) and test sets across 8
  languages\footnote{DE, EN, ES FR, IT, JA, RU, ZH}. The usual setup
  contains $1000$ samples each or training and dev, and $4000$ for test, across $4$
  classes (topics). The aim was to create a class balanced sets (uniform class prior),
  which gives us $250$ samples per topic in both training and dev, and $1000$
  samples per topic in the test. However, not every language in the original Reuters Multilingual Corpus (RCV)
  has enough examples, hence the class prior is not uniform~\citep{Schwenk:2018:MLDoc}.
  Moreover, it only covers a small subset ($6000$ samples in total) of the actual RCV corpus,
  and results from such as smaller subset tend be less certain. To address this, we use the
  MLDoc data preparation scripts, and create $5$ different splits of the data, where each split contains the same aforementioned number of training, dev and test samples. This is analogous to a 5-fold cross-validation scheme. The mean and standard deviations across 5 splits are reported during evaluation. The experimental results show that such a robust evaluation is needed as the standard deviation across 5 splits is noticeable (see Section~\ref{sec:res} and Appendix E).

  IndicNLP-suite~\citep{Kakawani:2020:IndicNLP} contains several resources for NLP in Indian languages. From this suite, we take the IndicNLP news articles (INA for short) classification dataset, and prepare a cross-lingual setup similar to that of MLDoc. The INA comprises of $9$
  languages\footnote{BN, GU, KN, ML, MR, OR, PA, TE, TA} covering $7$ classes (topics). However, not all the $7$ topics are present in the news articles across all the 9 languages. In order to make cross-lingual experiments across multiple languages, we consider two setups: A two-class setup covering all $9$ languages, and a three-class setup covering $5$ languages. We keep at most of $250$ samples per topic in both training and dev, and $1000$ samples per topic in the test. Finally, we create $5$ such splits, which allows us to report mean and standard deviations.

  As we re-processed both MLDoc and INA datasets, we call the newer versions as MLDoc5x and INA5x respectively, where $5$ represents the five different splits. Details in Appendix B.
  \begin{table}[!ht]
    \centering
    \caption{In language classification accuracy (in \%) on the dev sets of MLDoc5x for various hyper-parameters of MBay-6L seed model. The embedding dimension is fixed to 256 and the classifier is GLCU. \label{tab:hyper_acc}}
    \scalebox{0.76}{
      \begin{tabular}{c|ccccccr} \toprule
        $\omega$ & EN    & DE    & FR    & IT    & ES    & RU   & Avg.  (s.d.)  \\
        \midrule
        5e-02   & 85.34 & 88.82 & 89.28 & 78.74 & 88.32 & 77.38 & 84.65  (4.84) \\
        5e-03   & 85.88 & 90.72 & \textbf{89.70} & \textbf{80.78} & \textbf{89.36} & \textbf{79.78} & \textbf{86.04}  (4.34) \\
        5e-04   & \textbf{86.50} & \textbf{90.88} & 89.68 & 79.88 & 88.62 & 79.34 & 85.82  (4.58) \\
        \bottomrule
      \end{tabular}
    }
  \end{table}
\subsection{MBay configurations}
\label{ssec:hyper}
The proposed Bayesian multilingual document model has two important
hyper-parameters, i.e.,
latent (embedding) dimension \(K\) and $\ell_2$ regularization weight \(\omega\)
corresponding  to the model parameters $\{\mb{T}\s\}\, \forall \, \ell$. We fixed the embedding dimension to 256 and explored $\omega \in \{5e-02, 5e-03, 5e-04\}$.
The prior distribution~\eqref{eq:prior} was set to $\mathcal{N}(\mb{w} \mid \mb{0}, (0.1)\mb{I})$ and the variational distribution $q(\mb{w})$ was initialized to be the same as prior. This enabled us to use same learning rate for both mean and variance parameters. The number of Monte Carlo samples $R$ for approximating the objective function \eqref{eq:objective} was set to $8$, which we found to be a reasonable trade-off between computation complexity and convergence speed. A maximum batch size of $4096$ was used during training. A
constant learning rate of $5e-02$ was used both during training and inference. The model is trained for a maximum of $100$ epochs and inference is done fora maximum of $50$ iterations. Our models are implemented using PyTorch~\citep{pytorch:2017} and will be made public.

%
\begin{table*}[!t]
  \centering
  \caption{Results on MLDoc5x. $a$: Averaging sentence embeddings. $g$: Results taken from official GitHub repository. $p$: Max-pooling over encoder outputs. $t$: Input trimmed to 128 tokens. $m:$ Input trimmed to maximum sequence length. ZS$^*$: Zero-shot. IL$^*$: In-language. \label{tab:mldoc5x}}
  \scalebox{0.9}{
    \begin{tabular}{ll|cccccccc|c|c} \toprule
      &  & \multicolumn{8}{c|}{{Zero-shot transfer (source language to the rest)}} & ZS$^*$ & IL$^*$\\
      Model & Classifier & EN & DE & FR & IT & ES & RU & JA & ZH &  Avg. & Avg. \\
      \midrule
      LASER$^{ag}$ & MLP & 73.28 & 73.47 & 71.98 & 70.84 & 68.13 & 69.08 & 66.29 & 72.53 & 70.70 & 88.46 \\
      \midrule
      LASER$^a$   & MLP  & 71.43 & 72.57 & 74.73 & 70.02 & 71.25 & 68.27 & 54.82 & 68.35 & 68.93 & \textbf{88.91} \\
      LASER$^a$   & LR   & 70.52 & 73.12 & 75.80 & 70.56 & 74.99 & 66.27 & 48.37 & 68.46 & 65.81 & 88.65 \\
      LASER$^p$   & MLP  & 74.76 & 75.02 & 75.93 & 69.55 & 69.41 & 69.32 & 60.95 & 68.00 & 70.37 & 87.81 \\
      LASER$^p$   & LR   & 73.97 & 75.19 & 75.75 & 70.22 & 73.93 & 68.68 & 61.70 & 69.34 & \textbf{71.10} & 87.87 \\
      \midrule
      XLM-R-stsb$^t$ & MLP & 74.29 & 72.48 & 74.02 & 70.84 & 70.24 & 69.08 & 72.06 & 70.28 & \textbf{71.66} & \textbf{87.09} \\
      XLM-R-stsb$^t$ & LR  & 72.87 & 70.87 & 72.61 & 68.49 & 68.31 & 65.76 & 70.75 & 69.47 & 69.89 & 85.78 \\
      XLM-R-stsb$^m$ & MLP & 68.11 & 68.10 & 69.80 & 66.22 & 65.92 & 66.98 & 64.03 & 63.94 & 66.64 & 85.80 \\
      XLM-R-stsb$^m$ & LR  & 67.18 & 67.79 & 68.10 & 64.47 & 64.17 & 64.35 & 63.17 & 62.17 & 65.17 & 84.63 \\
      \midrule
      Distil-mUSE$^t$ & MLP & 75.92 & 74.86 & 75.90 & 72.51 & 74.01 & 69.84 & 69.77 & 71.40 & 73.03 & 88.14 \\
      Distil-mUSE$^t$ & LR  & 77.02 & 76.41 & 76.98 & 76.04 & 74.80 & 71.28 & 72.02 & 74.08 & \textbf{74.83} & \textbf{88.33} \\
      Distil-mUSE$^m$ & MLP & 73.34 & 73.33 & 73.81 & 71.72 & 74.66 & 69.85 & 68.05 & 71.29 & 72.01 & 87.92 \\
      Distil-mUSE$^m$ & LR  & 74.55 & 75.30 & 75.57 & 74.17 & 74.57 & 71.07 & 68.51 & 73.75 & 73.44 & 88.14 \\
      \midrule
      LaBSE$^t$ & MLP & 80.02 & 79.29 & 79.11 & 78.70 & 79.93 & 77.16 & 78.42 & 76.90 & 78.69 & \textbf{89.93} \\
      LaBSE$^t$ & LR  & 80.48 & 79.91 & 80.00 & 79.08 & 80.02 & 76.71 & 78.60 & 78.04 & \textbf{79.13} & 89.85 \\
      LaBSE$^m$ & MLP & 79.29 & 80.07 & 80.36 & 78.76 & 79.07 & 76.05 & 78.49 & 76.17 & 78.53 & 89.84 \\
      LaBSE$^m$ & LR  & 80.27 & 80.22 & 79.81 & 78.72 & 79.66 & 75.92 & 78.95 & 77.47 & 78.88 & 89.77 \\
      \midrule
      MBay   & GLC  & 65.04 & 64.71 & 65.39 & 61.65 & 62.28 & 57.05 & 54.02 & 59.99 & 61.27 & 83.70 \\
      MBay   & GLCU & 74.14 & 70.07 & 72.40 & 73.20 & 72.64 & 67.57 & 64.48 & 66.03 & 70.06 & 85.30 \\
      MBay & GLCU-P & 74.08 & 70.46 & 72.59 & 73.07 & 72.62 & 67.66 & 64.35 & 66.49 & \textbf{70.16} & 86.05 \\
      MBay  & LR    & 70.33 & 70.43 & 71.18 & 67.30 & 68.45 & 62.35 & 59.81 & 65.69 & 66.94 & \textbf{86.35} \\
      MBay  &  LRU  & 72.59 & 71.02 & 71.87 & 72.04 & 71.31 & 65.11 & 63.21 & 65.17 & 69.04 & 86.08 \\
      MBay  & LRU-P & 72.92 & 70.83 & 71.67 & 71.89 & 69.57 & 64.30 & 63.42 & 65.59 & 68.77 & 86.02 \\
      \bottomrule
  \end{tabular}}
  \normalsize
\end{table*}

\subsection{Topic ID systems for MBay}
\label{ssec:proposed}
In total we trained 4 different linear classifiers on the embeddings extracted from MBay model. The first two linear classifiers, GLC and LR are trained using only the point estimates of the embeddings, i.e., using only the mean parameter ($\bs{\nu}$). The next two classifiers, GLCU and LRU are trained with the full posterior distributions of embeddings, $q(\mb{w})=\mathcal{N}(\mb{w}\mid \bs{\nu}, \mathrm{diag}(\bs{\gamma})\I)$, as described in Section~\ref{sec:clf}. To better illustrate the importance of uncertainties during the test (prediction) time, we used the trained GLC and LR models, but during the prediction, we evaluate likelihood using the full posterior distributions (along with uncertainties) of the test document embeddings. This is valid because both GLC and GLCU have exactly the same model parameters (Section~\ref{ssec:clf1}). Similarly LR and LRU have exactly the same model parameters (Section~\ref{ssec:clf2}). We represent these two classifiers as GLCU-P and LRU-P, where -P denotes \textit{uncertainty exploited only during prediction}.

The generative classifiers (GLC, GLCU) have no
hyper-parameters to tune. We added $\ell_2$ regularization term with weight $\alpha \in \{1e-4, \ldots 5e+1\}$  for the parameters of LR, LRU. This classifier was trained for a maximum 100
epochs using \adam~with a constant learning rate of $5e-2$. For LRU, we used $M=32$ for the empirical
approximation~\eqref{eq:clf_softmax_mc}. $M> 32$ did not affect the
classification performance significantly but, lower values degraded the
performance about 5\%.

Initially three MBay models were trained on 6 languages (DE, EN, ES, FR, IT, RU) with different hyper-parameters. We performed in-language classification on MLDoc5x using GLCU on these 6 languages and picked the MBay model configuration that gave the best performance on dev set. These results are presented in Table~\ref{tab:hyper_acc}. We denote this seed model as MBay-6L. This model with the same hyper-parameter $(\omega=5e-03)$ is then extended independently to \{JA, ZH\}, and to 9 Indian languages using EN as pivot (bridge). More details are in Appendix C.


%


\subsection{Pre-trained multilingual models}
\label{ssec:pretrained}
There are numerous pre-trained multilingual models from which we picked the following\footnote{More details are given in Table 4 from Appendix E.}  based on their diversity in architecture, training criterion and overall performance.

\noindent \textbf{LASER}~\citep{Mikel:2019:MMS} is based on seq2seq BiLSTM trained in 223M parallel sentence covering 93 languages, sharing a common sub-word vocabulary. The language-agnostic embeddings are obtained by forward propagating through the encoder followed by a pooling layer.

\noindent \textbf{XLM-R-stsb}~\citep{Reimers:2020:Making} is based on sentence transformers~\citep{Reimers:2019:SBERT} and XLM-R~\citep{Alexis:2020:XLR}, where knowledge distillation is used to adapt the the multilingual student model XLM-R to align the representations from BERT.

\noindent \textbf{LaBSE}~\citep{Feng:2020:LaBSE} is based on dual-encoder architecture and is trained on 17B monolingual sentences for MLM, and on 6B translation pairs for translation ranking task, covering 109 languages. The pre-trained model is available for public, whereas the exact training data is not.

\noindent \textbf{Distill-mUSE} is multilingual knowledge distilled version of mUSE~\citep{Yang:2020:mUSE}. While the original was trained on 15 languages, this version supports 50 languages~\citep{Reimers:2020:Making}.

We trained two different classifiers on the embeddings extracted pre-trained multilingual language models. The first one is a two layer perceptron (MLP) widely used in prior works~\citep{Mikel:2019:MMS}. The second one is a logistic regression (LR).

%
%
\begin{table*}[t]
  \centering
  \caption{Results on INA5x 2-class setup.\label{tab:ina_2c}}
  \scalebox{0.9}{
    \begin{tabular}{ll|ccccccccc|c|c} \toprule
      &  & \multicolumn{9}{c|}{\textsc{Zero-shot transfer (from language)}} & ZS$^*$ & IL$^*$\\
      Model & CLF. & BN & GU & KN & ML & MR & OR & PA & TA & TE &  Avg. & Avg. \\
      \midrule
      LASER$^p$   & MLP  & 76.41 & - & - & 77.26 & 77.19 &  - & - & 74.52 & 77.46 & \textbf{76.57} & 92.77 \\
      LASER$^p$   & LR   & 76.70 & - & - & 77.86 & 74.85 &  - & - & 74.22 & 77.15 & 76.16 & 90.90 \\
      \midrule
      XLM-R-stsb$^t$ & MLP & 90.77 & 89.38 & 94.13 & 91.92 & 91.78 & 92.29 & 89.88 & 91.95 & 92.72 & 91.65 & 95.52 \\
      XLM-R-stsb$^t$ & LR  & 90.84 & 88.82 & 93.32 & 92.40 & 90.78 & 91.43 & 89.00 & 91.88 & 91.03 & \textbf{91.08} & 96.09 \\
      XLM-R-stsb$^m$ & MLP & 85.60 & 88.57 & 91.95 & 89.42 & 87.58 & 91.39 & 88.62 & 86.75 & 90.37 & 88.92 & 95.66 \\
      XLM-R-stsb$^m$ & LR  & 87.69 & 87.68 & 92.74 & 90.98 & 87.61 & 91.83 & 88.61 & 85.41 & 90.99 & 89.28 & 95.48 \\ \midrule
      Distil-mUSE$^t$ & MLP & - & 84.80 & - & - & 82.14 & - & - & - & - & 83.47 & 93.65 \\
      Distil-mUSE$^t$ & LR  & - & 83.24 & - & - & 86.39 & - & - & - & - & \textbf{84.81} & 93.50 \\
      Distil-mUSE$^m$ & MLP & - & 77.65 & - & - & 72.53 & - & - & - & - & 75.09 & 92.98 \\
      Distil-mUSE$^m$ & LR  & - & 83.79 & - & - & 76.51 & - & - & - & - & 80.15 & 92.55 \\
      \midrule
      LaBSE$^t$ & MLP & 96.41 & 96.91 & 97.18 & 97.31 & 97.43 & 96.83 & 96.41 & 97.43 & 97.13 & 97.00 & 98.03 \\
      LaBSE$^t$ & LR  & 96.62 & 96.71 & 97.78 & 97.21 & 97.31 & 97.37 & 96.40 & 97.34 & 97.63 & 97.15 & 98.06 \\
      LaBSE$^m$ & MLP & 95.92 & 96.98 & 97.48 & 97.37 & 97.54 & 96.46 & 96.90 & 96.93 & 97.19 & 96.97 & 97.95 \\
      LaBSE$^m$ & LR  & 96.60 & 96.69 & 97.56 & 97.23 & 97.56 & 97.47 & 96.87 & 97.38 & 97.75 & \textbf{97.23} & 97.98 \\
      \midrule
      MBay   & GLC  & 67.22 & 50.86 & 82.83 & 50.47 & 50.71 & 82.20 & 49.54 & 83.64 & 85.38 & 66.98 & 73.48 \\
      MBay   & GLCU & 91.89 & 93.54 & 94.69 & 93.72 & 94.01 & 94.67 & 93.93 & 93.47 & 94.43 & 93.82 & 96.67 \\
      MBay & GLCU-P & 92.67 & 93.24 & 94.96 & 94.38 & 95.48 & 94.70 & 93.65 & 93.86 & 94.66 & \textbf{94.18} & 97.03 \\
      MBay   & LR   & 91.59 & 90.76 & 93.41 & 93.14 & 93.34 & 92.91 & 91.54 & 90.96 & 92.91 & 92.28 & 96.44 \\
      MBay   & LRU  & 92.50 & 91.95 & 94.49 & 93.56 & 94.67 & 94.17 & 92.27 & 92.16 & 94.24 & 93.34 & 96.80 \\
      MBay  & LRU-P & 92.36 & 91.97 & 94.45 & 93.37 & 94.58 & 93.26 & 91.34 & 92.28 & 94.09 & 93.08 & 96.67 \\
      \bottomrule
  \end{tabular}}
  \normalsize
\end{table*}
%
%
\begin{table*}[!ht]
  \centering
  \caption{Results on INA5x 3-class setup.\label{tab:ina_3c}}
  \scalebox{0.9}{
    \begin{tabular}{ll|ccccc|c|c} \toprule
      &  & \multicolumn{5}{c|}{\textsc{Zero-shot transfer (from language)}} & ZS$^*$ & IL$^*$\\
      Model & CLF. & GU & ML & OR & PA & TE &  Avg. & Avg. \\
      \midrule
      LASER$^p$   & MLP  & - & 72.83 & - & - & 83.90 & 78.37 & 93.51 \\
      LASER$^p$   & LR   & - & 73.97 & - & - & 83.39 & \textbf{78.68} & 93.38 \\
      \midrule
      XLM-R-stsb$^t$ & MLP & 90.57 & 91.91 & 91.79 & 89.43 & 93.05 & \textbf{91.35} & 95.22 \\
      XLM-R-stsb$^t$ & LR  & 88.94 & 90.99 & 90.68 & 86.41 & 91.38 & 89.68 & 93.79 \\
      XLM-R-stsb$^m$ & MLP & 86.99 & 87.86 & 90.16 & 87.23 & 90.09 & 88.47 & 95.39 \\
      XLM-R-stsb$^m$ & LR  & 82.62 & 87.95 & 88.77 & 83.78 & 90.12 & 86.65 & 93.78 \\
      \midrule
      LaBSE$^t$ & MLP & 97.43 & 97.41 & 97.11 & 96.52 & 97.14 & 97.12 & 98.21 \\
      LaBSE$^t$ & LR  & 95.91 & 96.86 & 96.69 & 95.28 & 97.01 & 96.35 & 97.64\\
      LaBSE$^m$ & MLP & 97.45 & 97.47 & 96.65 & 96.92 & 97.18 & \textbf{97.13} & 98.09 \\
      LaBSE$^m$ & LR  & 95.62 & 96.79 & 96.59 & 95.25 & 97.00 & 96.25 & 97.47 \\
      \midrule
      MBay   & GLC  & 32.40 & 33.87 & 82.01 & 32.68 & 84.89 & 53.17 & 57.82 \\
      MBay   & GLCU & 89.21 & 89.92 & 91.12 & 88.63 & 90.58 & 89.89 & 95.29 \\
      MBay & GLCU-P & 89.42 & 90.89 & 91.19 & 88.97 & 90.97 & \textbf{90.29} & 95.78 \\
      MBay   & LR   & 87.16 & 87.18 & 89.18 & 85.84 & 88.84 & 87.64 & 94.95 \\
      MBay   & LRU  & 87.87 & 89.91 & 89.84 & 86.64 & 90.22 & 88.90 & 95.26 \\
      MBay  & LRU-P & 87.00 & 90.09 & 89.92 & 86.76 & 90.22 & 88.80 & 95.16 \\
      \bottomrule
  \end{tabular}}
\end{table*}
%
%
\section{Results and discussion}
\label{sec:res}
Here we present only the main zero-shot transfer results, while the detailed results are given in the Appendix E (Tables 5, 6 and 7). The mean and std. deviation across 5 splits for MLDoc and INA are only presented in the Appendix. For LASER + MLP system, we observed around 14 points of std. deviation across 5 splits in MLDoc5x when transferring from IT$\rightarrow$DE (Table 5, Appendix E). Higher ($>5$) std. deviations are also observed for other pre-trained models in different transfer directions. This suggests that one needs to have a robust evaluation scheme in order to study and compare the performance of multilingual models across various languages and tasks. Further, when reporting average results, care should be taken to separate them into in-language vs (zero-shot) transfer directions. A simple way to summarize the results is to compute average only across transfer directions for every language (excluding the source language). This gives us an idea of how well the model can transfer to other languages on an average. The in-language classification accuracy across various languages should be reported separately.

The first row from Table~\ref{tab:mldoc5x} show the results with LASER on the original single MLDoc split. We tried to replicate the results, but observed significant variance for JA and ZH (see Table 3 in Appendix D).
All the subsequent rows are the average results on MLDoc5x for systems based on various pre-trained models and the proposed MBay model. For most of the pre-trained models we can see that LR performs slightly better than MLP in zero-shot transfer setting. The results from MBay are comparable to LASER and XLM-R-stsb, while LaBSE outperforms all the other systems. Moreover, in case of MBay, we can see that generative classifier exploiting uncertainty outperforms the discriminative classifiers. This suggests that in the common embedding space, our classifiers are able to exploit the estimated uncertainty from the MBay model.

The Tables~\ref{tab:ina_2c} and \ref{tab:ina_3c} show average results in INA5x under 2-class and 3-class settings respectively. Here we can see that MBay outperforms other pre-trained models except LaBSE; and XLM-R-stsb in 3-class setting. The poor performance of LASER and XLM-R could be attributed to less and low-quality training data for these (mid-resource) Indian languages. The objective function of mUSE and LaBSE are similar where as the quality and quantity of the training data is much different. LaBSE was trained on large amounts of high-quality (manually verified, and filtered) data, which could explain its superior performance. Unfortunately, the exact training data used for LaBSE is not available for public.


 \subsection{Unsupervised topic discovery}
 \label{sec:topic_dis}
 To further understand our multilingual model, we took the point estimates (mean parameter, $\bs{\nu}$) of document embeddings of 5 languages from test set of MLDoc corpus, and clustered them using $k$-means with 10 clusters. We took cluster centroids ($\bar{\mb{c}}_k$) of the 4 most dense clusters and projected these vectors on to the individual language specific subspaces $\{\mb{T}\s\}$.
 \begin{equation}
   \bs{\theta}\s_k = \mb{T}\s \bar{\mb{c}}_k\,, \quad \forall \ell=1\dots5, \,\, \forall k=1\dots4
 \end{equation}
 The magnitude of values in $\bs{\theta}\s_k \in \mathbb{R}^{V\s}$ indicates the significance (representativeness) of the words from language $\ell$ to the cluster $k$. Table~\ref{tab:clusters} presents top 4 words from each language for each of the 4 clusters. Note that we did not use any parallel dictionary in our model, yet we can discover semantically related words across multiple languages.

 \begin{table}[ht]
    \centering
   \setlength{\tabcolsep}{2.5pt}
  \caption{Top 4 representative words from each language for top 4 dense clusters obtained via $k$-means. \label{tab:clusters}}
   \scalebox{1}{
   \begin{tabular}{l|l} \toprule
     EN  &  resellers, dealer, stabilises, volatility \\
     DE  &  uberschussen, marktpreise, preislich, anzukommen \\
     FR  &  negociants, volatilite, nourrie, commercialisent \\
     IT  &  responsabilizzati, concessionari, volatilita, compra \\
     ES  &  subprimes, pingues, mora, abastecer \\
     \midrule
     EN  &  tyrant, gorostiaga, authoritarianism, tribal \\
     DE  &  friedlicher, friedliebenden, kriegsverbrecher, \\
        &  anfuhrern \\
     FR  &  colonel, pacifiquement, gorostiaga, tyran \\
     IT  &  pacifiste, sradicato, miloseviæ, tribali \\
     ES  &  tirano, vil, magrebi, tribales \\
     \midrule
     EN  &  inflation, inflationary, predictions, slowdown \\
     DE  &  wirtschaftsindikatoren, haushaltsdefiziten, \\
         &  inflationsrate, wirtschaftsdaten \\
     FR  &  inflationniste, inflation, inflationnistes, pronostics \\
     IT  &  inflazione, inflazionistici, inflazionistiche, ciclica \\
     ES  &  inflacion, inflacionistas, predicciones, coyuntural \\
     \midrule
     EN  &  overvaluation, yen, lira, dollar \\
     DE  &  dollars, yuan, wechselkurses, chinesischem \\
     FR  &  surevaluation, croissent, dollar, degonflement \\
     IT  &  sopravvalutazione, valutari, yen, dollaro \\
     ES  &  dolar, fly, yen, redondeo \\
     \bottomrule
     \end{tabular}
   }
 \end{table}

\section{Conclusions}
\label{sec:concl}
In this paper, we revisited zero-shot cross-lingual topic identification. We identified shortcomings in the evaluation protocol of MLDoc corpus. We proposed a simple robust alternative by creating 5 different splits and reporting the mean and standard deviation of the results. The same protocol was extended to Indic news articles dataset covering 9 languages. We benchmarked some of the diverse and popular pre-trained models on the new evaluation protocol covering 17 languages (128 transfer directions). We also presented a Bayesian multilingual document model, which learns language-independent document embeddings along with their uncertainties.
We propagated the uncertainties into a generative and discriminative linear
classifier for zero-shot cross-lingual topic ID. Our proposed system in budget friendly in terms of computation, while at the same time performs competitively to other large scale pre-trained models such as LASER, and XLM-R. We believe our MBay model can act as a strong baseline for future research works in the direction of cross-lingual topic ID. We observe that  there is a need for creating a larger and diverse dataset covering several topics and languages.

\section{Limitations}
While we aimed to cover 17 languages, the number of topics in classification experiments are at most 4. There is a need to benchmark these systems on a diverse and large multi-label cross-lingual dataset. The proposed MBay model is build on bag-of-words simplification and may not be a suitable choice for fine-grained semantic similarity tasks.

%

\pagebreak

\newpage
\bibliography{refs}

\newpage

\appendix

\section*{Appendix}

The supplementary material provides more information on the statistics of the data used. We also present the detailed results (mean and standard deviation) for all the models across all the datasets.

\section{Data for multilingual training}
\label{app:data_stats}
The data statistics are with reference to Section~\ref{ssec:data_prep}.
\begin{table*}[!ht]
	\centering
	\caption{Statistics of the data used in training and extending the MBay model. Sentences and tokens are in millions (M).\label{tab:data_stats}}
	\small{
		\begin{tabular}{llccrrr} \toprule
			Group & Language & ISO code  & Parallel pairs & Sentences (M) & Tokens (M) & Vocabulary size \\
			\midrule
			$\mathcal{E}$ & English   & EN & $\mathcal{E} \cup \mathcal{U} \cup \mathcal{I}\,\setminus\,$\{KN\} & 3.89 & 154.46 & 100k \\
			$\mathcal{E}$ & French    & FR & $\mathcal{E}$   & 1.62 & 73.38  & 100k \\
			$\mathcal{E}$ & German    & DE & $\mathcal{E}$   & 0.85 & 33.02  & 100k\\
			$\mathcal{E}$ & Italian   & IT & $\mathcal{E}$   & 0.67 & 25.18  & 100k\\
			$\mathcal{E}$ & Russian   & RU & $\mathcal{E}$   & 1.11 & 37.01  & 100k\\
			$\mathcal{E}$ & Spanish   & ES & $\mathcal{E}$   & 1.64 & 74.03  & 100k\\
			$\mathcal{U}$ & Chinese    & ZH &  \{EN, JA\}     & 1.19 & 54.84  & 100k\\
			$\mathcal{U}$ & Japanese  & JA & \{EN, ZH\}      & 0.37 & 21.15  & 100k\\
			$\mathcal{I}$ & Kannada   & KN & $\mathcal{I}$   & 0.36 & 7.92   & 25521 \\
			$\mathcal{I}$ & Bengali   & BN & $\mathcal{I}\, \cup$ \{EN\}  & 0.95 & 20.05  & 36925 \\
			$\mathcal{I}$ & Gujarati  & GU & $\mathcal{I}\, \cup$ \{EN\}  & 0.75 & 15.92  & 28268 \\
			$\mathcal{I}$ & Malayalam & ML & $\mathcal{I}\, \cup$ \{EN\}  & 0.57 & 13.89  & 36877 \\
			$\mathcal{I}$ & Marathi   & MR & $\mathcal{I}\, \cup$ \{EN\}  & 0.86 & 18.43  & 30557 \\
			$\mathcal{I}$ & Odia      & OR & $\mathcal{I}\, \cup$ \{EN\}  & 0.50 & 11.36  & 25450 \\
			$\mathcal{I}$ & Punjabi   & PA & $\mathcal{I}\, \cup$ \{EN\}  & 0.95 & 15.34  & 24209 \\
			$\mathcal{I}$ & Tamil     & TA & $\mathcal{I}\, \cup$ \{EN\}  & 0.93 & 21.21  & 33960 \\
			$\mathcal{I}$ & Telugu    & TE & $\mathcal{I}\, \cup$ \{EN\}  & 0.68 & 11.96  & 32548 \\
			\midrule
			& Total  & & & 17.89 & 609.16 & \\
			\bottomrule
		\end{tabular}
	}
\end{table*}

\begin{itemize}
	\item We considered only top 450k sentence from EN-JA pair from  CCAligned corpus, which was further filtered based on heuristics, resulting in 185k parallel sentences.
	\item From UNPC(v1), we considered only top 2 million sentences.
	\item The initial seed model (MBay-6L) was trained on 6 languages (DE, EN, ES, FR, IT, RU) using the data from Europarl, UNPC, MultiUN, Global-Voices, and News-Commentary. From these datasets, we considered only those sentences that are at least 30 words long.
	\item The seed model (MBay-6L) is extended to JA and ZH languages with the help of parallel data from UNPC, MultiUN, Wikipedia (EN-JA)\footnote{\url{https://alaginrc.nict.go.jp/WikiCorpus/index_E.html}}, filtered CCAligned (EN-JA), Global-Voices and News-Commentary.
	\item The seed model (MBay-6L) is extended to 9 Indian languages (BN, GU, ML, MR, KN, OR, PA, TA, TE) with the help of parallel data from CVIT (PIB, MKB), Samanantar (indic2indic), Global-Voices and News-Commentary datasets. From these datasets, we considered only those sentences that are at least 10 words long.
	\item The Table~\ref{tab:data_stats} shows the detailed statistics of the number of sentences and their parallel languages across all the 17 languages.
\end{itemize}

\section{Data for Topic ID}
\label{app:data_topic_id}
This section presents the statistics of MLDoc5x and INA5x topic ID datasets created for the experiments reported in this paper. We attempted to keep about 250 examples per topic in each training and development sets, and 1000 examples per topic in the test set. We created 5 such splits and the average number of examples per language-set-topic are illustrated in Tables~\ref{tab:ina5x_data_stats}. The original data for languages GU, ML, PA, were smaller, hence they have smaller number of examples per set.
\begin{table*}[!ht]
	\centering
	\caption{Number of examples in each \textit{topic} for every \textit{language} in RCV (MLDoc5x) and IndicNLP news articles (INA5x) datasets. Under each topic, the three columns represent training, development and test \textit{sets} respectively. Each number represents the average number (rounded to nearest integer) of examples across 5 splits for the respective \textit{language-set-topic}. }
	\label{tab:ina5x_data_stats}
	\small{
		\begin{tabular}{l|rrr|rrr|rrr|rrr}
			\toprule
			\multirow{2}{*}{\textbf{Lang.}} & \multicolumn{12}{c}{\textbf{MLDoc5x  Topics}} \\
			\cmidrule{2-13}
			& \multicolumn{3}{c|}{CCAT} & \multicolumn{3}{c|}{ECAT} & \multicolumn{3}{c|}{GCAT} & \multicolumn{3}{c}{MCAT} \\
			\midrule
			DE & 257 & 270 & 1019 & 245 & 259 & 936 & 251 & 259 &1022 & 247 & 237 & 1023 \\
			EN & 257 & 270 & 1019 & 245 & 259 & 936 & 251 & 259 &1022 & 247 & 237 & 1023 \\
			ES & 305 & 310 & 1186 & 198 & 197 & 782 & 205 & 214 & 816 & 292 & 279 & 1216 \\
			FR & 257 & 270 & 1019 & 245 & 259 & 936 & 251 & 259 &1022 & 247 & 237 & 1023 \\
			IT &  257 & 270 & 1019 & 245 & 259 & 936 & 251 & 259 &1022 & 247 & 237 & 1023 \\
			JA & 257 & 270 & 1019 & 245 & 259 & 936 & 251 & 259 &1022 & 247 & 237 & 1023 \\
			RU & 274 & 283 & 1081 & 255 & 265 & 1023 & 204 & 256 & 819 & 267 & 256 & 1077 \\
			ZH & 306 & 313 & 1193 & 282 & 312 & 1187 & 118 & 93 & 401 & 294 & 282 & 1219 \\
			\midrule
			& \multicolumn{9}{c}{\textbf{INA5x Topics}} \\
			\cmidrule{2-13}
			& \multicolumn{3}{c|}{Entertainment} & \multicolumn{3}{c|}{Sports} & \multicolumn{3}{c|}{Business} \\
			\midrule
			BN & 250 & 250 & 1001 & 250 & 249 & 999 & -   & -   & -\\
			GU & 34  & 37  & 149  & 37  & 36  & 147 & 40  & 38  & 160 \\
			KN & 233 & 237 & 931  & 235 & 231 & 938 & -   & -   & -\\
			ML & 78  & 81  & 319  & 83  & 80  & 328 & 75  & 76  & 302 \\
			MR & 121 & 122 & 494  & 120 & 123 & 494 & -   & -   & - \\
			OR & 237 & 232 & 928  & 236 & 233 & 948 & 237 & 237 & 959 \\
			PA & 40  & 41  & 163  & 40  & 39  & 165 & 43  & 44  & 170 \\
			TA & 236 & 232 & 928  & 231 & 235 & 935 & -   & -   & - \\
			TE & 248 & 249 & 988  & 252 & 253 & 997 & 250 & 247 & 1015 \\
			\bottomrule
		\end{tabular}
	}
\end{table*}

\section{MBay models}
\label{app:mbay_systems}
\begin{itemize}
	\item The initial (seed) MBay models were trained on 6 languages (DE, EN, ES, FR, IT, RU) using the parallel data (7.48M sentences) described in Appendix~\ref{app:data_stats}. The training took about 25 hrs on a single NVIDIA RTX A6000 with 48 GB of memory. The trained model has 154M parameters. This model trained on 6 languages is referred as MBay-6L.
	\item The MBay-6L seed model was extended to JA, ZH using EN as pivot. It was trained on 3.3M parallel sentences, and took about 11 hrs on a similar GPU. This extended training added 51.4M additional parameters for JA and ZH. During training, the parameters of EN were frozen and the parameters of other languages (DE, ES, FR, IT, RU) were not loaded as they are not required.
	\item The MBay-6L seed model was extended to 9 Indian languages using EN as pivot. It was trained on 7.29M parallel sentences, and took about 21hrs to train on a similar GPU. This added 96.2M additional parameters to represent 9 Indian languages. As the vocabulary sizes for these languages is not as big as other high-resources languages (Table~\ref{tab:data_stats}, the number of additional parameters were also relatively less.
\end{itemize}

\section{MLDoc results with LASER}
\label{app:laser_repl}

We tried to replicate the MLDoc results using LASER, however we found significant differences is few language directions.
\begin{table*}[!th]
	\def\arraystretch{1}
	\centering
	\caption{Discrepancy in replicating the results of LASER + MLP system.}
	\label{tab:laser_mldoc_discr}
	\small
	\begin{tabular}{c|rrrrrrrr} \toprule
		& \multicolumn{8}{c}{\textsc{Test language}} \\
		&  \EN & \DE & \FR & \IT & \ES & \RU & \JA & \ZH \\ \midrule
		\EN & -0.42 & -1.70 & -4.72 & -0.23 & 3.05 & 0.13 & -1.67 & -3.30 \\
		\DE & 0.93 & -0.68 & 0.65 & -2.18 & -1.83 & 0.30 & \textbf{-7.17} & -0.78 \\
		\FR & -1.60 & -2.40 & -0.58 & -2.22 & -1.98 & -0.17 & \textbf{-6.87} &
		\textbf{-13.31} \\
		\IT & -2.38 & -2.59 & -2.25 & 2.78 & 3.20 & \textbf{5.86} & \textbf{-6.47} &
		\textbf{-10.72} \\
		\ES & -0.24 & -2.70 & -2.85 & -1.58 & -4.93 & \textbf{7.80} & -4.05 &
		\textbf{8.95} \\
		\RU & 0.53 & -2.76 & 0.35 & 2.55 & 2.25 & -1.00 & -2.05 & 2.17 \\
		\JA & \textbf{10.70} & \textbf{14.37} & \textbf{10.28} & \textbf{6.23} &
		\textbf{11.87} & \textbf{9.53} & -0.07 & \textbf{15.32} \\
		\ZH & 2.02 & 0.20 & 1.01 & 0.38 & \textbf{5.85} & 0.17 & 4.15 & 0.51 \\
		\bottomrule
	\end{tabular}
\end{table*}

The Table~\ref{tab:laser_mldoc_discr} shows the absolute differences in the results we obtained as compared the ones reported in the official github repository: \texttt{\url{https://github.com/facebookresearch/LASER/tree/main/tasks/mldoc}}. In Table~\ref{tab:laser_mldoc_discr}, a positive value indicates that we obtained a better result, while a negative value indicates the opposite.



\section{Detailed results}
\label{app:detailed}

\begin{table*}[!ht]
	\centering
	\caption{Pre-trained models and their download URL.} \label{tab:pretrained}
	\scalebox{0.8}{
		\begin{tabular}{ll}
			\toprule
			Model & URL \\
			\midrule
			LASER & \texttt{https://github.com/facebookresearch/LASER} \\
			XLM-R-stsb & \texttt{https://huggingface.co/sentence-transformers/stsb-xlm-r-multilingual} \\
			LaBSE & \texttt{https://huggingface.co/sentence-transformers/LaBSE} \\
			Distil-mUSE & \texttt{https://huggingface.co/sentence-transformers/distiluse-base-multilingual-cased-v2} \\
			\bottomrule
		\end{tabular}
	}
\end{table*}

Here we present the detailed results i.e., mean and std.dev. across 5-splits in all the transfer directions. For each pre-trained multilingual model, we only show the results of the system that yielded best downstream performance. Notice in Tables~\ref{tab:mldoc_detailed}, \ref{tab:ina2x_detailed} and \ref{tab:ina3x_detailed} that the high std.dev. indicates that the choosing a different training / dev / test split could result in different performance of the system. The original MLDoc is sampled from RCV multilingual corpus and had only one such split and hence couldn't capture the variance in the results. The first two parts in Table~\ref{tab:mldoc_detailed} show LASER$^p$ + MLP and LASER$^p$ + LR. Notice that for LR the variance across 5 splits is much lower as compared to MLP.

\begin{table*}[!ht]
	\def\arraystretch{1}
	\centering
	\scalebox{0.85}{
		\begin{tabular}{c|llllllll} \toprule
			& \multicolumn{8}{c}{\textsc{Test language}} \\
			&  EN & DE & FR & IT & ES & RU & JA & ZH \\
			\midrule
			& \multicolumn{8}{c}{LASER$^p$ + MLP} \\
			\midrule
			EN & \cellcolor{lightgray}{87.0 (0.7)} & 85.5 (1.4) & 83.0 (3.8) & 68.4 (4.6) & 77.8 (2.2) & 67.4 (2.7) & 67.1 (3.8) & 74.1 (0.7) \\
			DE &  73.7 (4.9) &\cellcolor{lightgray}{92.1 (0.1)} & 83.9 (0.7) & 73.4 (1.4) & 81.0 (1.9) & 66.9 (5.7) & 71.6 (4.9) &\textbf{74.6 (9.5)} \\
			FR &  76.2 (0.5) & 88.0 (0.4) &\cellcolor{lightgray}{90.5 (0.5)} & 72.3 (1.2) & 79.9 (2.0) & 68.0 (0.9) & 69.8 (0.5) & 77.2 (1.0) \\
			IT & \textbf{61.3 (12.5)} &\textbf{78.8 (14.1)} &\textbf{77.1 (8.3)} &\cellcolor{lightgray}{84.2 (0.7)} & 76.0 (2.6) & 63.4 (4.6) &\textbf{61.2 (7.6)} &\textbf{69.1 (6.5)} \\
			ES &  64.4 (0.3) & 81.9 (1.9) & 78.8 (3.1) & 74.4 (2.5) &\cellcolor{lightgray}{92.5 (0.1)} & 57.7 (3.3) &\textbf{64.1 (7.6)} &\textbf{64.5 (12.2)} \\
			RU &  64.9 (2.3) & 78.1 (5.5) &\textbf{70.2 (7.3)} & 66.7 (3.1) & 70.2 (5.4) &\cellcolor{lightgray}{83.4 (0.3)} &\textbf{67.5 (6.3)} &\textbf{67.5 (11.3)} \\
			JA &  58.6 (1.1) & 70.7 (3.5) & 62.8 (2.7) & 57.3 (1.3) & 59.8 (1.9) & 51.7 (2.9) &\cellcolor{lightgray}{85.9 (0.1)} &\textbf{65.8 (8.1)} \\
			ZH & \textbf{63.7 (7.9)} & 76.3 (5.2) &\textbf{70.5 (7.9)} &\textbf{64.8 (6.7)} & 68.5 (3.9) & 61.0 (1.9) & 71.2 (2.5) &\cellcolor{lightgray}{86.9 (0.7)} \\
			\midrule
			& \multicolumn{8}{c}{LASER$^p$ + LR} \\
			\midrule
			EN & \cellcolor{lightgray}{87.3 (0.5)} & 86.2 (1.1) & 81.9 (1.2) & 67.2 (2.1) & 76.8 (2.0) & 66.0 (2.0) & 66.3 (2.3) & 73.5 (2.1) \\
			DE &  73.0 (1.4) &\cellcolor{lightgray}{92.4 (0.3)} & 83.0 (0.8) & 73.5 (1.1) & 81.1 (1.2) & 68.0 (0.8) & 71.3 (0.9) & 76.4 (2.0) \\
			FR &  75.1 (0.3) & 88.2 (1.0) &\cellcolor{lightgray}{90.3 (0.5)} & 72.8 (1.1) & 79.3 (1.0) & 67.9 (2.0) & 69.3 (1.9) & 77.6 (2.4) \\
			IT &  61.1 (1.8) & 80.4 (2.6) & 77.1 (2.0) &\cellcolor{lightgray}{84.5 (0.7)} & 76.6 (1.7) & 64.4 (1.0) &\textbf{61.9 (3.6)} &\textbf{70.1 (3.5)} \\
			ES &  68.5 (0.5) & 84.1 (0.8) & 81.6 (1.3) & 76.0 (1.3) &\cellcolor{lightgray}{92.7 (0.4)} & 64.9 (1.2) & 69.5 (1.5) & 72.9 (2.3) \\
			RU &  64.8 (0.8) & 76.9 (1.4) & 69.3 (2.4) & 66.6 (1.3) & 69.1 (1.8) &\cellcolor{lightgray}{83.1 (0.4)} & 67.0 (1.2) &\textbf{66.9 (3.6)} \\
			JA &  60.3 (1.4) & 72.9 (1.1) & 64.1 (1.9) & 56.4 (1.5) & 60.4 (1.6) & 50.1 (1.2) &\cellcolor{lightgray}{85.7 (0.4)} & 67.6 (1.7) \\
			ZH &  64.2 (2.4) & 77.5 (1.2) & 71.6 (2.2) & 66.8 (0.9) & 68.4 (1.5) & 64.0 (1.7) & 73.1 (0.4) &\cellcolor{lightgray}{87.0 (0.7)} \\
			\midrule
			& \multicolumn{8}{c}{XLM-R-stsb$^t$ + MLP} \\
			\midrule
			EN & \cellcolor{lightgray}{88.0 (0.7)} & 85.1 (1.2) & 79.4 (1.7) & 69.4 (1.0) & 78.8 (0.8) &\textbf{66.3 (3.7)} & 68.8 (1.6) & 72.2 (2.7) \\
			DE &  75.1 (0.7) &\cellcolor{lightgray}{92.5 (0.4)} & 83.0 (0.5) & 71.5 (2.4) & 77.9 (0.9) & 61.7 (0.5) & 69.5 (2.6) &\textbf{68.7 (4.1)} \\
			FR &  77.3 (1.2) & 87.8 (1.4) &\cellcolor{lightgray}{89.7 (0.8)} & 72.9 (2.0) & 78.6 (2.2) & 63.7 (2.7) & 69.0 (2.9) &\textbf{68.9 (3.3)} \\
			IT &  69.0 (1.2) & 82.6 (1.2) & 79.1 (1.6) &\cellcolor{lightgray}{83.0 (0.6)} & 77.9 (1.7) &\textbf{56.5 (3.1)} & 67.6 (1.8) & 63.2 (1.7) \\
			ES &  71.3 (1.5) & 78.7 (2.6) & 78.4 (1.8) & 72.4 (2.6) &\cellcolor{lightgray}{92.3 (0.3)} &\textbf{57.8 (6.1)} & 68.4 (1.5) & 64.7 (2.1) \\
			RU &  69.3 (0.7) & 76.9 (2.8) & 75.3 (1.2) & 65.0 (2.8) & 70.9 (2.1) &\cellcolor{lightgray}{82.9 (0.2)} & 63.6 (1.6) &\textbf{62.4 (3.8)} \\
			JA &  71.6 (1.4) & 83.0 (0.7) & 77.1 (1.2) & 66.4 (1.6) & 73.7 (1.0) &\textbf{61.7 (4.1)} &\cellcolor{lightgray}{83.7 (0.7)} & 70.9 (1.2) \\
			ZH &  70.1 (2.8) & 79.4 (2.7) & 75.1 (2.5) & 64.3 (2.5) &\textbf{69.3 (3.1)} & 62.6 (2.2) & 71.2 (1.3) &\cellcolor{lightgray}{84.7 (0.4)} \\
			\midrule
			& \multicolumn{8}{c}{Distil-mUSE$^t$ + LR} \\
			\midrule
			EN & \cellcolor{lightgray}{89.3 (0.3)} & 85.9 (1.4) & 82.4 (1.3) & 69.1 (1.9) & 77.7 (1.5) & 62.3 (2.4) &\textbf{65.2 (3.3)} & 79.3 (1.3) \\
			DE &  77.7 (1.0) &\cellcolor{lightgray}{93.1 (0.4)} & 84.9 (0.3) & 73.0 (0.6) & 79.6 (1.5) & 66.4 (1.0) &\textbf{65.5 (4.2)} & 80.1 (1.0) \\
			FR &  78.7 (0.6) & 89.4 (0.4) &\cellcolor{lightgray}{90.7 (0.4)} & 73.1 (1.0) & 80.0 (1.7) & 64.4 (1.7) & 63.3 (1.1) & 80.1 (1.3) \\
			IT &  72.1 (2.2) & 83.4 (1.2) & 81.4 (0.9) &\cellcolor{lightgray}{83.7 (0.6)} & 79.8 (1.9) & 64.5 (1.8) &\textbf{60.7 (3.7)} & 77.2 (1.1) \\
			ES &  77.7 (1.1) & 85.7 (1.3) & 82.5 (1.1) & 75.4 (0.7) &\cellcolor{lightgray}{92.7 (0.4)} &\textbf{63.9 (3.1)} & 60.8 (1.8) & 76.1 (1.8) \\
			RU &  70.8 (2.6) & 82.0 (1.9) & 74.3 (2.3) & 66.5 (1.0) & 68.8 (2.4) &\cellcolor{lightgray}{83.2 (0.4)} & 64.1 (0.8) & 71.1 (2.1) \\
			JA &  70.4 (2.1) & 76.6 (1.9) & 71.9 (1.9) & 62.7 (1.3) & 66.3 (2.7) & 56.0 (2.4) &\cellcolor{lightgray}{85.0 (0.6)} & 75.6 (1.3) \\
			ZH &  76.1 (2.2) & 83.6 (2.1) & 77.9 (2.1) & 71.2 (2.0) & 75.5 (1.6) & 67.0 (2.2) & 65.0 (2.2) &\cellcolor{lightgray}{87.4 (0.4)} \\
			\midrule
			& \multicolumn{8}{c}{LaBSE$^t$ + LR} \\
			\midrule
			EN & \cellcolor{lightgray}{90.6 (0.4)} & 89.0 (0.9) & 87.8 (0.6) & 76.2 (1.3) & 82.7 (1.0) &\textbf{69.6 (3.4)} & 75.8 (1.1) & 82.2 (0.6) \\
			DE &  77.8 (1.5) &\cellcolor{lightgray}{93.8 (0.3)} & 88.2 (0.7) & 76.2 (0.8) & 84.9 (2.2) & 72.0 (2.9) & 76.5 (1.5) & 83.7 (0.7) \\
			FR &  81.4 (0.4) & 91.1 (0.6) &\cellcolor{lightgray}{92.1 (0.4)} & 76.3 (1.2) & 83.6 (1.4) &\textbf{71.0 (3.2)} & 73.7 (1.1) & 82.9 (1.3) \\
			IT &  73.3 (1.2) & 87.0 (0.6) & 84.2 (1.1) &\cellcolor{lightgray}{86.6 (0.2)} & 85.2 (0.7) & 71.6 (1.1) & 71.7 (1.2) & 80.6 (1.3) \\
			ES &  77.9 (1.5) & 89.9 (0.6) & 86.9 (1.3) & 81.0 (0.9) &\cellcolor{lightgray}{93.9 (0.3)} &\textbf{68.9 (3.0)} & 75.4 (1.5) & 81.6 (1.0) \\
			RU &  73.3 (1.1) & 86.1 (1.8) & 80.7 (2.2) & 74.1 (1.3) &\textbf{72.8 (3.4)} &\cellcolor{lightgray}{86.0 (0.5)} & 71.2 (1.8) & 78.8 (1.5) \\
			JA &  76.8 (0.5) & 87.1 (1.3) & 83.7 (1.4) & 72.8 (0.5) & 79.5 (1.1) & 67.6 (2.7) &\cellcolor{lightgray}{86.2 (0.5)} & 82.8 (1.8) \\
			ZH &  76.8 (0.6) & 86.6 (2.4) & 83.4 (2.7) & 74.2 (2.5) & 79.2 (1.5) &\textbf{69.4 (3.8)} & 76.7 (1.9) &\cellcolor{lightgray}{89.6 (0.6)} \\
			\midrule
			& \multicolumn{8}{c}{MBay + GLCU-P} \\
			\midrule
			EN & \cellcolor{lightgray}{86.8 (0.3)} & 85.6 (0.3) & 82.4 (1.0) & 70.3 (1.1) & 78.5 (0.6) & 65.0 (1.7) & 66.8 (1.6) & 70.1 (1.5) \\
			DE &  75.2 (0.9) &\cellcolor{lightgray}{91.1 (0.5)} & 85.6 (0.5) & 70.6 (1.0) & 79.8 (1.1) & 54.9 (1.7) & 57.9 (2.7) & 69.3 (1.3) \\
			FR &  75.3 (0.6) & 87.0 (0.7) &\cellcolor{lightgray}{89.8 (0.4)} & 74.1 (0.4) & 80.9 (0.6) & 65.7 (1.3) & 51.9 (1.1) & 73.2 (1.3) \\
			IT &  73.8 (1.0) & 84.6 (0.9) & 83.9 (0.5) &\cellcolor{lightgray}{82.1 (0.7)} & 82.9 (0.9) & 58.3 (2.4) & 59.5 (1.8) & 68.4 (0.9) \\
			ES &  74.1 (0.6) & 84.6 (1.1) & 82.5 (0.7) & 75.3 (0.7) &\cellcolor{lightgray}{89.1 (0.2)} & 58.5 (1.8) & 64.7 (1.6) & 68.6 (1.2) \\
			RU &  66.8 (1.7) & 74.9 (1.6) & 75.3 (1.7) & 67.6 (1.3) & 73.6 (1.2) &\cellcolor{lightgray}{81.2 (0.5)} & 55.3 (1.5) & 60.1 (2.8) \\
			JA &  67.5 (0.9) & 76.6 (1.2) & 68.1 (1.5) & 59.9 (0.8) & 67.4 (0.8) & 51.6 (1.5) &\cellcolor{lightgray}{84.7 (0.7)} & 59.4 (1.2) \\
			ZH &  69.1 (0.9) & 77.9 (1.8) & 73.5 (1.9) & 63.5 (1.2) & 66.9 (1.5) & 50.8 (1.9) & 63.7 (1.1) &\cellcolor{lightgray}{83.7 (0.6)} \\
			\bottomrule
		\end{tabular}
	}
	\caption{Detailed classification results on the MLDoc5x test sets using various models with best downstream classification performance. Values in the parenthesis indicate the std.dev. across 5 splits. Bold values indicate the numbers with std.dev $> 3$. $p$: Max-pooling over encoder outputs. $t$: Input trimmed to 128 tokens. $m:$ Input trimmed to maximum sequence length.}
	\label{tab:mldoc_detailed}
\end{table*}

\begin{table*}[!ht]
	\centering
	\small{
		\begin{tabular}{c|lllllllll} \toprule
			& \multicolumn{9}{c}{\textsc{Test language}} \\
			&  BN & GU & KN & ML & MR & OR & PA & TA & TE \\
			\midrule
			& \multicolumn{9}{c}{LASER$^t$ + MLP} \\
			\midrule
			BN & \cellcolor{lightgray}{95.4 (0.5)} & - & - & 90.8 (0.5) & 85.4 (2.5) & - & - & \textbf{57.4 (8.1)} &\textbf{72.0 (5.5)} \\
			ML &  86.2 (1.6) & - & - & \cellcolor{lightgray}{93.4 (1.8)} & 86.3 (2.8) & - & - & \textbf{63.7 (5.2)} & 72.8 (2.2) \\
			MR & \textbf{89.2 (3.0)} & - & - & 88.8 (2.4) &\cellcolor{lightgray}{93.4 (1.0)} & - & - &\textbf{59.8 (7.0)} & 70.9 (2.3) \\
			TA &  80.6 (2.3) & - & - & 77.8 (2.2) &\textbf{76.9 (5.1)} & - & - & \cellcolor{lightgray}{88.1 (0.7)} &\textbf{62.8 (5.1)} \\
			TE & \textbf{83.6 (3.9)} & - & - &\textbf{83.9 (3.3)} & 86.0 (1.9) & - & - & 56.3 (2.9) &\cellcolor{lightgray}{93.7 (0.5)} \\
			\midrule
			& \multicolumn{9}{c}{XLM-R-stsb$^t$ + LR} \\
			\midrule
			BN & \cellcolor{lightgray}{95.0 (0.3)} & 92.4 (0.9) & 87.4 (2.2) & 93.6 (1.1) & 93.6 (1.2) & 91.3 (1.7) &\textbf{91.5 (3.0)} & 93.7 (1.4) &\textbf{83.2 (3.1)} \\
			GU &  84.7 (1.9) &\cellcolor{lightgray}{93.4 (2.0)} & 84.0 (1.7) & 91.5 (1.0) & 92.7 (1.2) &\textbf{90.6 (3.5)} &\textbf{89.3 (4.0)} & 91.9 (1.4) & 85.8 (1.9) \\
			KN & \textbf{88.5 (3.3)} & 94.3 (0.7) &\cellcolor{lightgray}{93.7 (0.7)} & 95.6 (0.6) & 95.0 (0.5) & 93.0 (1.0) & 94.1 (1.6) & 96.6 (0.3) & 89.5 (1.1) \\
			ML &  88.3 (2.4) & 93.6 (1.2) & 89.6 (2.1) &\cellcolor{lightgray}{95.3 (0.9)} & 95.6 (0.8) & 94.5 (0.5) & 94.8 (0.7) & 95.6 (1.0) & 87.2 (0.5) \\
			MR &  86.7 (2.1) & 94.8 (0.7) & 83.9 (2.3) & 94.4 (0.7) &\cellcolor{lightgray}{96.4 (0.3)} & 94.5 (0.7) & 93.8 (2.0) & 94.1 (0.4) & 84.1 (1.3) \\
			OR & \textbf{90.1 (3.4)} & 93.4 (1.8) & 86.6 (1.4) & 93.0 (2.9) & 95.3 (1.0) &\cellcolor{lightgray}{96.3 (0.4)} & 92.8 (2.8) & 94.1 (1.5) & 86.0 (2.5) \\
			PA &  87.3 (1.9) & 91.9 (1.3) & 84.8 (1.3) & 90.7 (2.0) & 91.7 (1.9) & 91.7 (1.4) &\cellcolor{lightgray}{94.5 (2.0)} & 92.7 (1.8) & 81.3 (2.3) \\
			TA &  90.6 (2.0) & 92.9 (2.0) & 89.7 (1.4) & 95.0 (1.0) & 95.6 (0.6) &\textbf{90.0 (3.4)} & 91.5 (2.4) &\cellcolor{lightgray}{97.9 (0.4)} & 89.7 (0.3) \\
			TE &  87.2 (2.6) & 91.9 (1.5) & 89.0 (1.8) & 92.0 (2.6) & 93.5 (2.0) &\textbf{89.8 (4.4)} &\textbf{90.1 (3.7)} & 94.8 (1.9) &\cellcolor{lightgray}{94.7 (0.7)} \\
			\midrule
			& \multicolumn{9}{c}{LaBSE$^m$ + LR} \\
			\midrule
			BN & \cellcolor{lightgray}{97.0 (0.4)} & 95.5 (2.1) & 95.0 (0.7) & 97.3 (0.8) & 96.1 (1.7) & 97.6 (0.4) & 97.6 (1.2) & 97.5 (2.2) & 96.2 (1.1) \\
			GU &  95.4 (0.7) &\cellcolor{lightgray}{96.9 (1.2)} & 95.6 (0.5) & 97.3 (0.3) & 96.5 (0.6) & 97.3 (0.5) & 97.4 (1.0) & 97.2 (1.0) & 96.7 (0.4) \\
			KN &  95.7 (1.1) & 96.7 (0.8) &\cellcolor{lightgray}{96.9 (0.2)} & 98.8 (0.6) & 97.9 (0.3) & 97.3 (0.4) & 98.1 (0.4) & 98.6 (0.3) & 97.5 (0.6) \\
			ML &  95.2 (1.1) & 97.0 (1.1) & 96.2 (0.3) &\cellcolor{lightgray}{98.4 (0.3)} & 98.0 (0.3) & 97.6 (0.6) & 97.8 (0.7) & 98.6 (0.4) & 97.4 (0.4) \\
			MR &  96.5 (0.3) & 97.2 (1.0) & 96.5 (0.2) & 98.4 (0.6) &\cellcolor{lightgray}{98.1 (0.2)} & 98.1 (0.2) & 98.4 (0.5) & 98.6 (0.2) & 97.0 (0.3) \\
			OR &  96.5 (0.6) & 96.6 (0.7) & 96.2 (0.2) & 98.0 (0.6) & 98.0 (0.2) &\cellcolor{lightgray}{98.4 (0.2)} & 98.5 (0.8) & 98.6 (0.4) & 97.4 (0.3) \\
			PA &  96.4 (0.3) & 95.6 (0.9) & 95.5 (0.2) & 97.7 (0.4) & 97.5 (0.3) & 97.9 (0.2) &\cellcolor{lightgray}{98.5 (0.5)} & 98.1 (0.4) & 96.3 (0.3) \\
			TA &  96.6 (0.4) & 96.5 (1.2) & 96.5 (0.3) & 98.1 (0.2) & 97.6 (0.6) & 98.0 (0.2) & 98.0 (0.9) &\cellcolor{lightgray}{99.2 (0.1)} & 97.7 (0.2) \\
			TE &  96.7 (0.4) & 97.4 (0.9) & 96.9 (0.3) & 98.3 (0.5) & 97.9 (0.2) & 97.4 (0.5) & 98.3 (0.9) & 99.1 (0.2) &\cellcolor{lightgray}{98.5 (0.2)} \\
			\midrule
			& \multicolumn{9}{c}{MBay + GLCU-P} \\
			\midrule
			BN & \cellcolor{lightgray}{96.2 (0.6)} & 91.8 (2.4) & 91.5 (0.7) & 90.0 (0.8) & 94.5 (0.4) & 93.6 (0.4) & 91.8 (1.1) & 96.4 (0.5) & 92.0 (0.9) \\
			GU &  94.4 (0.9) &\cellcolor{lightgray}{97.0 (0.9)} & 91.6 (0.7) & 92.0 (0.3) & 94.3 (1.0) & 94.8 (0.5) & 91.2 (0.9) & 96.7 (0.6) & 90.8 (1.6) \\
			KN &  91.4 (1.4) & 94.4 (1.9) &\cellcolor{lightgray}{95.6 (0.5)} & 94.6 (1.1) & 95.9 (0.4) & 95.8 (0.6) & 95.1 (1.3) & 98.0 (0.3) & 94.5 (0.5) \\
			ML &  93.1 (1.9) & 94.0 (2.3) & 93.5 (0.5) &\cellcolor{lightgray}{97.2 (0.6)} & 94.9 (1.2) & 94.6 (1.6) & 93.8 (2.3) & 98.0 (0.4) & 93.2 (0.3) \\
			MR &  94.4 (1.2) & 95.0 (1.4) & 94.3 (0.5) & 96.0 (0.3) &\cellcolor{lightgray}{96.7 (0.3)} & 96.5 (0.5) & 96.1 (1.1) & 98.2 (0.4) & 93.4 (0.7) \\
			OR &  93.9 (0.7) & 95.0 (0.9) & 93.1 (0.6) & 95.8 (0.7) & 96.5 (0.6) &\cellcolor{lightgray}{97.9 (0.2)} & 94.5 (1.4) & 96.9 (0.3) & 91.8 (0.5) \\
			PA & \textbf{92.2 (3.6)} & 93.4 (1.9) & 92.2 (0.8) & 93.9 (0.8) & 95.1 (0.6) & 94.0 (0.9) &\cellcolor{lightgray}{97.2 (0.8)} & 97.1 (0.6) & 91.2 (0.6) \\
			TA &  93.4 (0.4) & 94.3 (1.4) & 93.0 (0.8) & 94.4 (1.4) & 94.8 (0.9) & 93.5 (0.4) & 93.4 (0.7) &\cellcolor{lightgray}{98.7 (0.1)} & 94.1 (0.5) \\
			TE &  91.9 (1.2) & 95.3 (0.7) & 94.2 (0.6) & 94.2 (1.1) & 95.4 (0.6) & 94.7 (0.3) & 93.4 (1.6) & 98.1 (0.2) &\cellcolor{lightgray}{96.7 (0.4)} \\
			\bottomrule
		\end{tabular}
		\caption{Detailed classification results on the 2-class setup from INA5x test sets using various models with best downstream classification performance. Values in the parenthesis indicate the std.dev. across 5 splits. Bold values indicate the numbers with std.dev $> 3$. $p$: Max-pooling over encoder outputs. $t$: Input trimmed to 128 tokens. $m:$ Input trimmed to maximum sequence length.}
		\label{tab:ina2x_detailed}
	}
\end{table*}
\begin{table*}[!ht]
	\small{
		\centering
		\begin{tabular}{c|lllll} \toprule
			& \multicolumn{5}{c}{\textsc{Test language}} \\
			&  GU & ML & OR & PA & TE \\
			\midrule
			& \multicolumn{5}{c}{XLM-R-stsb$^t$ + MLP} \\
			\midrule
			GU & \cellcolor{lightgray}{94.2 (1.5)} & 92.6 (1.1) & 92.6 (1.0) & 91.3 (1.8) & 85.8 (2.9) \\
			ML &  93.2 (1.5) &\cellcolor{lightgray}{95.2 (0.7)} & 93.8 (0.5) & 94.0 (1.6) & 86.7 (1.8) \\
			OR &  94.0 (1.1) & 94.3 (0.9) &\cellcolor{lightgray}{96.4 (0.4)} & 93.1 (2.6) & 85.8 (2.0) \\
			PA &  91.6 (1.6) & 91.5 (1.4) & 92.3 (2.4) &\cellcolor{lightgray}{95.3 (1.2)} & 82.3 (2.9) \\
			TE &  93.5 (0.9) & 93.4 (1.0) & 92.9 (1.8) & 92.3 (1.0) &\cellcolor{lightgray}{94.9 (0.7)} \\
			\midrule
			& \multicolumn{5}{c}{LaBSE$^m$ + MLP} \\
			\midrule
			GU & \cellcolor{lightgray}{97.0 (1.0)} & 97.7 (0.4) & 97.6 (0.3) & 97.6 (1.1) & 97.0 (0.3) \\
			ML &  97.1 (0.8) &\cellcolor{lightgray}{98.3 (0.4)} & 97.5 (0.8) & 97.8 (0.6) & 97.4 (0.4) \\
			OR &  95.4 (1.7) & 97.5 (0.3) &\cellcolor{lightgray}{98.3 (0.3)} & 97.4 (0.6) & 96.3 (1.1) \\
			PA &  95.9 (0.9) & 97.4 (0.3) & 97.9 (0.3) &\cellcolor{lightgray}{98.5 (0.5)} & 96.4 (0.2) \\
			TE &  97.0 (1.0) & 97.8 (0.5) & 96.8 (0.9) & 97.1 (1.2) &\cellcolor{lightgray}{98.4 (0.3)} \\
			\midrule
			& \multicolumn{5}{c}{MBay + GLCU-P} \\
			\midrule
			GU & \cellcolor{lightgray}{96.0 (0.3)} & 90.6 (0.9) & 88.4 (0.9) & 89.2 (1.6) & 89.5 (1.3) \\
			ML &  92.6 (1.6) &\cellcolor{lightgray}{94.9 (0.7)} & 88.9 (1.3) & 91.3 (1.4) & 90.8 (0.6) \\
			OR &  91.6 (0.5) & 92.3 (0.6) &\cellcolor{lightgray}{96.5 (0.3)} & 91.0 (1.2) & 89.8 (0.7) \\
			PA &  91.1 (1.3) & 90.2 (1.1) & 85.0 (2.0) &\cellcolor{lightgray}{95.3 (1.1)} & 89.5 (0.7) \\
			TE &  93.2 (1.3) & 90.9 (0.6) & 87.5 (0.7) & 92.3 (1.4) &\cellcolor{lightgray}{96.1 (0.3)} \\
			\bottomrule
		\end{tabular}
		\caption{Detailed classification results on the 2-class setup from INA5x test sets using various models with best downstream classification performance. Values in the parenthesis indicate the std.dev. across 5 splits. Bold values indicate the numbers with std.dev $> 3$. $p$: Max-pooling over encoder outputs. $t$: Input trimmed to 128 tokens. $m:$ Input trimmed to maximum sequence length.}
		\label{tab:ina3x_detailed}
	}
\end{table*}

\end{document}